\documentclass{article}
\usepackage{ijcai25}

\usepackage{times}
\usepackage{soul}
\usepackage{url}
\usepackage[hidelinks]{hyperref}
\usepackage[utf8]{inputenc}
\usepackage[small]{caption}
\usepackage{graphicx}
\usepackage{amsmath}
\usepackage{amsthm}
\usepackage{booktabs}
\usepackage{algorithm}
\usepackage{algorithmic}
\usepackage[switch]{lineno}
\usepackage{multirow}
\usepackage{makecell}
\usepackage{arydshln}
\usepackage{enumitem}
\usepackage{xcolor}
\newcommand{\saleh}[1]

\title{Achieving Upper Bound Accuracy of Joint Training in Continual Learning}

\author{Saleh Momeni~~ and~~
Bing Liu
\affiliations
Department of Computer Science, University of Illinois Chicago, USA\\
\emails
smomen3@uic.edu,
liub@uic.edu
}

\begin{document}
\maketitle

\begin{abstract}
Continual learning has been an active research area in machine learning, focusing on incrementally learning a sequence of tasks. A key challenge is \textit{catastrophic forgetting} (CF), and most research efforts have been directed toward mitigating this issue. However, a significant gap remains between the accuracy achieved by state-of-the-art continual learning algorithms and the ideal or upper-bound accuracy achieved by training all tasks together jointly. This gap has hindered or even prevented the adoption of continual learning in applications, as accuracy is often of paramount importance. Recently, another challenge, termed \textit{inter-task class separation} (ICS), was also identified, which spurred a theoretical study into principled approaches for solving continual learning. Further research has shown that by leveraging the theory and the power of large foundation models, it is now possible to achieve upper-bound accuracy, which has been empirically validated using both text and image classification datasets. Continual learning is now ready for real-life applications. This paper surveys the main research leading to this achievement, justifies the approach both intuitively and from neuroscience research, and discusses insights gained.  
\end{abstract}

\section{Introduction}

The ability to continuously learn and accumulate knowledge over a lifetime is a hallmark of human intelligence. Yet, this critical capability is still absent from current dominant machine learning paradigms. Continual learning is the field of study that aims to provide AI agents with this capability. It incrementally learns a sequence of tasks. 
In supervised learning, a task refers to a set of classes learned together, while incremental learning means when training on a new task, the model has no access to the training data from previous tasks.

Continual learning has three main settings \cite{van2019three}: \textit{task incremental learning} (TIL), \textit{class incremental learning} (CIL), and \textit{domain incremental learning} (DIL). In TIL, each task is independent of other tasks. At test time, a task identity (task-id) is required for each instance to select the corresponding model for that task. DIL is a special case of TIL where each task shares the same set of classes but comes from a different domain. In CIL, a single model is trained to recognize all encountered classes, and no task-id is provided at test time.

A key challenge of continual learning is \textit{\textbf{catastrophic forgetting}} (\textbf{CF}), which refers to the phenomenon where, in learning a new task, the system must modify the network parameters, causing the model’s performance on earlier tasks to deteriorate \cite{McCloskey1989}. \textit{\textbf{Knowledge transfer}} (\textbf{KT}), which aims to leverage past knowledge to improve learning on new tasks, is also highly desirable \cite{ke2022survey}. 
KT is primarily applicable in TIL and DIL as their tasks may be from similar domains and can help each other. 

The CF problem has been solved for TIL as several approaches can prevent forgetting entirely \cite{serra2018overcoming,wortsman2020supermasks}. The upper-bound performance of TIL is that of multitask learning (MTL), which learns all tasks simultaneously. Several TIL approaches can already achieve the performance of MTL \cite{ke2021achieving}

This paper focuses on CIL. In addition to CF, recent work has identified another challenge for CIL known as \textit{\textbf{inter-task class separation}} (ICS) \cite{kim2022theoretical}. The ICS challenge arises from the difficulty of learning decision boundaries between new and previously learned classes without access to past data. The upper-bound performance of CIL is set by \textbf{joint training}, where all classes are learned simultaneously in a traditional supervised learning setup. Although many CIL methods have been proposed \cite{wang2024comprehensive}, due to CF and ICS challenges, they still have significant performance deterioration as more tasks are learned. Consequently, there is still a \textbf{major performance gap} between the accuracy of state-of-the-art CIL methods and joint training. This hinders the practical adoption of CIL methods. We are unaware of any report on the practical applications of CIL. 

In \cite{kim2022theoretical}, a theoretical study was conducted, outlining the key requirements a CIL method must meet to produce strong performance. Over the past years, several methods have been introduced based on the theory, demonstrating significant improvements compared to earlier approaches. More importantly, \cite{momeni2025continual} recently proposed a novel method that can finally achieve upper-bound accuracy for both text classification and image classification tasks, which has never been reported before. This method uses a \textit{self-supervised} \textbf{foundation model} only for feature extraction, keeping its parameters completely frozen. Empirical results reported in this study validate the methodology for achieving the upper-bound accuracy in CIL. This paper surveys the research leading to this achievement. The ability to achieve upper-bound accuracy shows that CIL is now ready for prime-time real-life applications.

The fact that CIL can reach its upper-bound performance using only a frozen foundation model raises important questions: ``\textit{is this approach justified?}'', 
``\textit{should a continual learning system learn new representations at all}?'', ``\textit{what happens if a foundation model does not have sufficiently rich features to represent a class of objects}?'', and ``\textit{does the human brain--often considered a large-scale `foundation model' or `world model'--learn features when encountering unfamiliar objects?}''. We discuss these questions in Section~\ref{sec.justifications}.

\section{Existing Approaches \& Their Limitations}
Continual learning has been widely explored, with most methods aimed at reducing CF. In this section, we review the main existing approaches and analyze their limitations, explaining why they fail to fully address CF or ICS, preventing them from reaching the upper-bound accuracy.

\subsection{Using Regularization}
Regularization-based methods mitigate CF by penalizing changes to important parameters learned for previous tasks through an additional regularization term in the loss function. These methods often require storing a frozen copy of the old model for reference and are typically categorized into \textit{\textbf{parameter-based}} and \textit{\textbf{distillation-based}} methods based on the target of the regularization.

Parameter-based methods constrain updates to parameters that are critical for previous tasks. A common approach to estimating importance is the Fisher Information Matrix \cite{kirkpatrick2017overcoming}, which quantifies how much each parameter contributes to past predictions. Various studies have explored more effective importance measures to improve stability and reduce forgetting \cite{aljundi2018memory}. 

Distillation-based methods preserve knowledge by aligning the current model’s predictions with those of previous tasks \cite{li2017learning}. This is typically achieved through a teacher-student framework, where the frozen old model acts as a teacher and the current model as a student \cite{hinton2015distilling}. Several works extend this idea by introducing improved teacher-student architectures and enhanced regularization strategies for better knowledge retention \cite{rebuffi2017icarl,buzzega2020dark}. 

Although these regularization-based approaches can reduce CF, they cannot eliminate CF because parameters from previous tasks still undergo changes despite the regularization constraints. They also do not address the ICS problem in CIL, as training focuses solely on establishing decision boundaries for the classes in the current task. Lacking mechanisms to manage these problems, the regularization-based approaches to CIL typically yield weak performance, falling significantly short of the upper-bound accuracy.

\subsection{Using Replay Data}
Replay-based methods mitigate CF by storing a subset of past task samples in a memory buffer and training on both the new and replayed data \cite{chaudhry2018efficient,huang2021continual,wang2022foster}. %
Earlier works commonly used reservoir sampling, where a fixed number of samples were uniformly obtained from training batches \cite{chaudhry2019tiny}. However, due to the limited memory, effective sample selection and retrieval are crucial for maintaining performance. More advanced buffer management techniques prioritize samples based on factors such as preserving latent decision boundaries, or identifying samples most negatively impacted by parameter updates \cite{shim2021online}. 
Additionally, some works improve storage efficiency through compression techniques or data augmentation to maximize the utility of stored examples \cite{bang2021rainbow}. 

The effectiveness of replay methods for CIL primarily depends on the size of the replay memory and how it is managed. In the extreme case where all past data is retained, the approach is equivalent to joint training, where new tasks are learned alongside all previously seen data. However, in practice, only a small subset can be stored, which is insufficient to fully adjust the past knowledge, leading to performance degradation. Since retaining large amounts of data contradicts the principles of continual learning, there has been a gradual shift away from replay-based methods.

An alternative approach to saving raw data-based replay is \textit{\textbf{pseudo-replay}}, where actual replay samples are replaced with synthetic data that resembles previously encountered samples. This approach is closely tied to continual learning of generative models, as the model must retain its ability to generate realistic past samples while learning new distributions \cite{shin2017continual}. 
Pseudo-replay is particularly common in the text domain, as language models inherently generate text, allowing them to serve as both task learners and data generators for pseudo-replay \cite{sun2020lamol,shao2023class}.

The success of this approach relies on the quality of the data generator, which ideally produces a sufficient amount of representative pseudo-replay data to mimic joint training. However, generating highly representative samples given limited training data is challenging, and the incremental training of the generator itself is prone to CF. Additionally, generating large amounts of pseudo-replay data is computationally expensive, making this method slow and impractical for achieving upper-bound accuracy of joint training.

\subsection{Gradient Projection}
Gradient projection aims to explicitly control parameter updates across tasks to mitigate CF. The primary technique within this category is \textit{\textbf{orthogonal projection}}, which modifies gradients to minimize interference with previously learned tasks. This is typically done by identifying subspaces that capture important gradient directions from past tasks and then projecting new gradients onto directions orthogonal to these subspaces to reduce interference \cite{farajtabar2020orthogonal,zeng2019continual}. 
Some studies further refine this approach by modifying previously learned tasks when gradient updates exhibit a positive correlation, allowing for selective retention of useful information \cite{lin2022beyond}.

Another notable technique is \textit{\textbf{null-space projection}}, where parameter updates are restricted to lie within the null space of previous task gradients \cite{kao2021natural}. 
This is achieved by computing the feature covariance matrix of past tasks and approximating its null space using singular value decomposition \cite{wang2021training}. 

These methods are primarily designed for TIL and have demonstrated effectiveness in mitigating CF. However, they often struggle with complex architectures, limiting their scalability and performance. For instance, the performance of orthogonal projection method in \cite{lin2022beyond} 
is significantly lower than that of the parameter isolation methods like \cite{kim2022theoretical}. Additionally, gradient orthogonality does not imply weight or parameter orthogonality, meaning that interference between tasks can still occur. These approaches also do not effectively address the ICS issue, thus limiting their overall effectiveness for CIL.

\subsection{Architectural Approaches}
\label{sec.architecture}
Architectural-based methods address CF by introducing structural changes to the model as new tasks are learned. These methods include techniques such as \textit{\textbf{network expansion}} and \textit{\textbf{parameter isolation}}.

Network expansion methods allocate independent subnetworks for each task by expanding the network with additional parameters, enabling task differentiation without interference. Some approaches introduce a new module for each task while enabling knowledge transfer through specialized connections \cite{rusu2016progressive}, 
whereas others employ mechanisms such as mixtures of experts or parallel branching to dynamically determine which modules should contribute to a given input \cite{aljundi2017expert}. 

In contrast, parameter isolation methods operate within a fixed model architecture by allocating isolated parameter subspaces for each task. These methods optimize binary masks to determine which neurons or parameters should be dedicated to a specific task while keeping previously allocated parameters frozen to prevent CF \cite{serra2018overcoming,wortsman2020supermasks}. 
Since the total network capacity remains constant, these methods often impose sparsity constraints to efficiently allocate parameters or selectively reuse existing ones \cite{gurbuz2022nispa}. 

Architectural-based methods usually de-correlate different tasks at the parameter level, which can almost avoid CF entirely \cite{serra2018overcoming}. However, a key limitation of these methods is their reliance on task identity to determine which subset of parameters should be used during inference. As a result, their applicability is largely confined to TIL settings, where task-ids are known.

To extend this approach to CIL, several studies have integrated TIL techniques with task identity predictors to identify the appropriate subnetwork for each input \cite{rajasegaran2020itaml,wang2024rehearsal}. 
Various task identity prediction methods have been explored, including learning a separate network specifically for task identification \cite{abati2020conditional}, using task-specific auto-encoders with reconstruction loss for task prediction \cite{aljundi2017expert}, and leveraging network expansion within an energy-based framework \cite{wang2022beef}. 
Moreover, task identity prediction has been shown to be closely related to out-of-distribution (OOD) detection \cite{kim2022theoretical}, making the integration of TIL methods with OOD detection a natural choice \cite{lin2024class}.

Combining a TIL framework with a task identity predictor can effectively prevent CF by fully decorrelating task-specific parameters. However, a key limitation is ICS. Since each subnetwork is trained independently on its respective task, it lacks awareness of global class boundaries, making task-id prediction difficult. Even with replay data, they still have difficulty achieving the accuracy of joint training \cite{kim2023learnability,lin2024class}. Additionally, the presence of multiple subnetworks results in varying feature representations for the same input depending on which task's parameters are used, further amplifying this problem. We will discuss this last point further in Section~\ref{sec.justifications}.

\subsection{Using Pre-trained Models}
Early continual learning techniques did not use any pre-trained models. With the advancement of pre-trained models in the past few years, recent techniques have increasingly leveraged pre-trained foundation models for improved accuracy. Several studies have shown that utilizing foundation models can benefit continual learning significantly from knowledge transfer to robustness to forgetting \cite{mehta2023empirical,kim2023learnability}. 

One widely adopted approach is \textit{\textbf{prompt learning}}, where the foundation model remains frozen, and a set of trainable prompts is learned to adapt the feature representations for the classes in continual learning tasks \cite{wang2022learning}. 
This allows the foundation model to adapt to different tasks while preserving its original knowledge. Various techniques have been proposed to enhance prompt learning, including task-specific prompts or weighted prompt aggregation through attention mechanisms \cite{smith2023coda,razdaibiedina2023progressive}. 
Although these methods keep the foundation model frozen, CF can still occur within the prompts in some cases. More importantly, they do not effectively address the ICS problem in CIL.

Another approach is to fine-tune the foundation model incrementally. Gradual fine-tuning of a pre-trained model has been shown to yield strong performance \cite{zhang2023slca}. Additionally, parameter-efficient transfer learning (PETL) with adapters allows for adaptation through minimal updates, reducing the risk of forgetting \cite{panos2023first}. 

In fact, a pre-trained foundation model can be incorporated into any of the previously discussed methods. While it improves performance, it does not eliminate fundamental limitations--if a method struggles with CF or ICS, using a foundation model does not resolve these issues, and thus there is still difficulty in achieving joint training accuracy of the same foundation model.

A recent promising approach is \textit{\textbf{prototype-based}} classification, where class prototypes are computed and stored in learning and later used in testing. In this approach, a foundation model is simply used for feature extraction without any parameter updates during learning. One well-known method in this category is the Nearest Class Mean (NCM) classifier, which assigns test samples to the closest class mean. Notably, this simple approach has been shown to outperform several strong baselines \cite{zhou2024continual}. This suggests that foundation models already provide high-quality representations for downstream tasks, and incremental updates may even degrade these representations due to CF. Prototype-based methods avoid CF entirely by eliminating parameter updates and address the ICS issue by maintaining class prototypes that establish natural decision boundaries between classes.

Other prototype-based approaches enhance decision boundaries by incorporating additional statistical information, such as the feature covariance matrix. Various methods have been explored, including Mahalanobis distance \cite{lee2018simple}, Linear Discriminant Analysis \cite{hayes2020lifelong}, and ridge regression \cite{zhuang2024gacl}. The key characteristic shared by these techniques is their reliance on data statistics rather than trainable parameters. 
Additionally, feature representations from foundation models can be improved through projection to a higher-dimensional space or kernel methods, which can improve class separability \cite{mcdonnell2024ranpac,momeni2025continual}. 

Notably, the KLDA method proposed in \cite{momeni2025continual}, despite having no learnable parameters, demonstrated that it can achieve the same accuracy as joint training --equivalent to fine-tuning the foundation model on all classes together. We provide a brief introduction to this method in Section~\ref{sec.prototype} and highlight key experimental results in Section~\ref{sec.evaluation}. 

\section{Applications}
Continual learning has been explored across various applications in computer vision and natural language processing (NLP). While the core challenge has been dealing with CF, different tasks introduce distinct challenges.

\subsection{Computer Vision}
Image classification has a main area of application for continual learning, where models must recognize an expanding set of categories over time \cite{wang2024comprehensive}. 
Object detection has also been explored, where the goal is to identify and localize multiple objects within an image. A key challenge in this area is that multiple objects from both old and new classes can appear together in the same image \cite{dong2021bridging}. 
Semantic segmentation extends classification to the pixel level, requiring models to assign labels to each pixel in an image \cite{michieli2021knowledge}. 
A major issue in this area is background shift, where previously learned objects may be mistakenly classified as background due to their absence in new tasks. Image generation has also been investigated in the context of continual learning, particularly in connection with pseudo-replay strategies \cite{shin2017continual}. 
Generative models must retain previously learned distributions while accommodating new visual styles, making them suitable for replay-based methods that regenerate past samples to mitigate forgetting.

\subsection{Natural Language Processing}
In NLP, continual learning has been widely studied in text classification \cite{shao2023class}. 
Another area of interest is slot filling, a key component of task-oriented dialogue systems. Here, models must extract relevant information, such as dates, locations, or product names, from user queries \cite{shen2019progressive}. 
Question answering presents additional challenges, as models must incorporate new knowledge sources while preserving previously acquired factual information \cite{yang2024continual}. 
Language acquisition has also been studied in the context of continual learning, where models must learn new linguistic structures, vocabulary, or even entirely new languages over time \cite{ke2022survey}. 

While continual learning spans a diverse range of applications in vision and NLP, this survey focuses specifically on classification tasks under the CIL setting.

\section{Problem Formulation \& Theoretical Results}
\label{sec.formulation}

In CIL, a model is trained sequentially on a series of tasks \(1, \dots, T\), where each task \( t \) has an input space \( \mathcal{X}^{(t)} \), a label space \( \mathcal{Y}^{(t)} \), and a training set \( \mathcal{D}^{(t)} = \{(\boldsymbol{x}_j^{(t)}, y_j^{(t)})\}_{j=1}^{n^{(t)}} \) drawn i.i.d. from \( \mathcal{P}_{\mathcal{X}^{(t)} \mathcal{Y}^{(t)}} \). The class labels across tasks are \textbf{disjoint}, i.e, \( \mathcal{Y}^{(i)} \cap \mathcal{Y}^{(k)} = \emptyset \) for all \( i \neq k \). The goal of CIL is to learn a unified model 
$
f: \bigcup_{t=1}^{T} \mathcal{X}^{(t)} \rightarrow \bigcup_{t=1}^{T} \mathcal{Y}^{(t)}
$ 
that can predict the class label for each test sample \( \boldsymbol{x} \) \textbf{without task-id}.  

With modest assumptions, \cite{kim2022theoretical} established that the CIL probability of predicting the $j$-th class label of task \( t \) for a test sample \( \boldsymbol{x} \) can be decomposed into two probabilities:  

\[
\mathbf{P}(y_j^{(t)} | \boldsymbol{x}) = \mathbf{P}(y_j^{(t)} | \boldsymbol{x}, t) \mathbf{P}(t | \boldsymbol{x}),
\]
where \( \mathbf{P}(y_j^{(t)} | \boldsymbol{x}, t) \) represents the \textbf{within-task prediction} (WP) probability, which is the probability of assigning \( \boldsymbol{x} \) to \( y_j^{(t)} \) given the task-id is known, and \( \mathbf{P}(t | \boldsymbol{x}) \) represents the \textbf{task-id prediction} (TP) probability, which estimates the likelihood that \( \boldsymbol{x} \) belongs to task \( t \).

The theoretical results in \cite{kim2022theoretical} include: (1) accurate WP and TP are \textit{necessary} and \textit{sufficient} conditions for good CIL accuracy and (2) TP and OOD detection for each task bound each other or are correlated with each other. These results have been generalized to the open world CIL setting in~\cite{kim2024open}.

Building on this theory, several CIL methods have been proposed that leverage a TIL method, such as HAT \cite{serra2018overcoming}, to learn an OOD detection model rather than a traditional classifier and protect it from CF \cite{kim2022theoretical,lin2024class}. Note that the OOD detection model for each task can be used to estimate both WP and TP probabilities. These approaches have achieved state-of-the-art results among methods that learn feature representations for each task. However, they still cannot achieve the upper-bound accuracy due to their difficulty in addressing the ICS issue.

\section{Prototype-based Continual Learning}
\label{sec.prototype}
An alternative approach that avoids representation learning altogether is prototype-based classification, where class prototypes are incrementally calculated using features from a foundation model without modifying the parameters. In this framework, the same input consistently produces the same representation throughout the learning process, making comparisons between tasks straightforward.
A well-known approach is the statistical technique \textit{linear discriminant analysis} (LDA), which represents the data of each class as a multivariate Gaussian distribution with a shared \textit{covariance matrix} across all classes and an individual \textit{mean} per class. Under this assumption, the log-posterior for a class $m$ can be computed based on a linear function:
\[
\log P(y = m \mid \mathbf{x}) = \mathbf{x}^\top \mathbf{\Sigma}^{-1} \mathbf{\mu}_m - \frac{1}{2} \mathbf{\mu}_m^\top \mathbf{\Sigma}^{-1} \mathbf{\mu}_m + \text{constant}
\]
Here, $\mathbf{x}$ is the feature representation of the input sample from the foundation model, and the term ``constant'' refers to \( P(\mathbf{x}) \) in Bayes' rule, which is identical for all classes. The class means and shared covariance matrix can be accumulated incrementally. 
Note that this solution is derived using explicit mathematical formulas rather than gradient-based optimization, meaning it has no learnable parameters and, therefore, completely avoids CF. Additionally, it inherently mitigates the ICS problem, as different classes are represented by distinct distributions, preserving clear decision boundaries. 

While prototype-based methods are effective with foundation model representations, the features can be further improved to enhance class separation. 
Kernel Linear Discriminant Analysis (KLDA) proposed by \cite{momeni2025continual} enhances LDA by using kernel methods to map features into a higher-dimensional space. When using the Radial Basis Function (RBF) kernel, this mapping extends to an infinite-dimensional space, making the classes more linearly separable. In practice, the kernel function is approximated using Random Fourier Features (RFF) \cite{rahimi2007random}, yielding a feature transformation that eliminates the need to compute the full kernel matrix.

This work exemplifies that the frozen representations of a foundation model when exploited properly, can achieve performance on par with the joint training upper-bound without any learnable parameters.

\vspace{+2mm}
\noindent
\textbf{Theoretical Justification:} As discussed earlier, effective WP and OOD detection are necessary and sufficient for successful CIL (Section \ref{sec.formulation}). In the KLDA approach proposed by \cite{momeni2025continual}, each class is modeled as a Gaussian distribution, effectively treating each class as an independent task. Then, WP is always correct (i.e., WP probability is always 1) and the Gaussian distribution naturally functions as an OOD detector--test samples that deviate significantly from the distribution receive lower likelihood scores.

\section{Empirical Validation} 
\label{sec.evaluation}
This section provides an overview of the experimental setup and key results directly taken from \cite{momeni2025continual}. The study evaluates KLDA on both text and image classification tasks, with a primary focus on text classification, as language foundation models are more mature.

\subsection{Experimental Setup}
For text classification, four datasets are used: CLINC consists of 150 dialogue intent classes, partitioned into 10 disjoint tasks. Banking focuses on banking-related intents with 77 classes, divided into 7 tasks. DBpedia contains 70 classes derived from Wikipedia articles, split into 7 tasks. HWU includes 64 dialogue intent classes spanning 20 domains, organized into 8 tasks. The tasks are generated by randomly shuffling the classes, and multiple runs are performed to account for variability. For image classification, experiments are conducted on CIFAR10, CIFAR100, TinyImageNet with 200 classes, and Stanford Cars, which comprises 196 classes. On these datasets, evaluation is conducted exclusively against the Joint upper-bound; therefore, task splits are not required.

The experiments from \cite{momeni2025continual} compare KLDA against several baselines: Vanilla is a sequential fine-tuning approach with no forgetting mitigation. Regularization-based methods include EWC \cite{kirkpatrick2017overcoming}, which penalizes changes to important network parameters, and KD \cite{hinton2015distilling}, which employs knowledge distillation. Prompt learning methods, such as L2P \cite{wang2022learning}, adapt the model using trainable prompts. Pseudo-replay approaches, including LAMOL \cite{sun2020lamol} and VAG \cite{shao2023class}, are also considered. Among prototype-based baselines, NCM and LDA are used. Joint fine-tunes the foundation model on all classes simultaneously and serves as the upper-bound.

The main experiments in \cite{momeni2025continual} are based on a BART-base foundation model with an encoder-decoder architecture, chosen for its compatibility with generative objectives and pseudo-replay requirements of the baseline methods. The study also evaluates using some other language foundation models, including T5-3b and Mistral-7b. 


For image classification, \textbf{self-supervised} foundation models such as DINOv2-small and DINOv2-base were used to avoid \textbf{information leakage}--a common issue in supervised pre-training where some or all classes encountered in continual learning have already been learned during initial training. Additionally, supervised pre-training is not scalable, as manually labeling a large number of classes is impractical.

The primary evaluation metric is \textbf{Last Accuracy}, which measures classification accuracy after all tasks have been learned in a dataset.

\begin{table*}[ht]
\begin{center}
\scalebox{1}{
\begin{tabular}{l|cccc}
\hline
\textbf{Method} & \textbf{CLINC} (10-T) & \textbf{Banking} (7-T) & \textbf{DBpedia} (7-T) & \textbf{HWU} (8-T) \\
\hline
Joint & 95.33 $\pm$\scriptsize{0.04} & 91.36 $\pm$\scriptsize{0.32} & 94.83 $\pm$\scriptsize{0.16} & 88.60 $\pm$\scriptsize{0.29} \\
\hdashline
Vanilla & 42.06 $\pm$\scriptsize{1.53} & 31.80 $\pm$\scriptsize{1.20} & 43.45 $\pm$\scriptsize{2.54} & 30.95 $\pm$\scriptsize{3.37} \\
EWC & 45.73 $\pm$\scriptsize{0.46} & 38.40 $\pm$\scriptsize{2.70} & 44.99 $\pm$\scriptsize{2.90} & 34.01 $\pm$\scriptsize{3.46} \\
KD & 36.33 $\pm$\scriptsize{0.86} & 27.40 $\pm$\scriptsize{1.59} & 42.10 $\pm$\scriptsize{2.40} & 25.46 $\pm$\scriptsize{2.13} \\
L2P & 30.66 $\pm$\scriptsize{2.46} & 31.45 $\pm$\scriptsize{0.55} & 23.52 $\pm$\scriptsize{1.54} & 24.04 $\pm$\scriptsize{0.88} \\
LAMOL & 58.42 $\pm$\scriptsize{0.84} & 42.60 $\pm$\scriptsize{1.36} & 48.61 $\pm$\scriptsize{1.82} & 44.85 $\pm$\scriptsize{1.57} \\
VAG & 76.42 $\pm$\scriptsize{0.90} & 59.34 $\pm$\scriptsize{1.28} & 65.40 $\pm$\scriptsize{1.52} & 56.88 $\pm$\scriptsize{1.22} \\
\hdashline
NCM & 83.60 $\pm$\scriptsize{0.00} & 71.10 $\pm$\scriptsize{0.00} & 75.70 $\pm$\scriptsize{0.00} & 73.30 $\pm$\scriptsize{0.00} \\
LDA & 93.71 $\pm$\scriptsize{0.00} & 89.09 $\pm$\scriptsize{0.00} & 93.42 $\pm$\scriptsize{0.00} & 86.41 $\pm$\scriptsize{0.00} \\
\hline
KLDA & 95.90 $\pm$\scriptsize{0.68} & 92.23 $\pm$\scriptsize{0.32} & 94.13 $\pm$\scriptsize{0.32} & 87.27 $\pm$\scriptsize{1.39} \\
KLDA--E & \textbf{96.62 $\pm$\scriptsize{0.08}} & \textbf{93.03 $\pm$\scriptsize{0.06}} & \textbf{94.53 $\pm$\scriptsize{0.12}} & \textbf{89.78 $\pm$\scriptsize{0.09}} \\
\hline
\end{tabular}
}
\caption{Final accuracy (\%) on text classification datasets for various methods, using a BART-base model with no replay buffer. The number of tasks per dataset is noted in parentheses (\#-T). Joint is the upper-bound for CIL, as it learns all classes simultaneously.}
\vspace{-1mm}
\label{tab:main-results}
\end{center}
\end{table*}

\begin{table*}[t]
    \centering
    \begin{minipage}{0.495\textwidth}
        \centering
        \resizebox{\textwidth}{!}{
            \begin{tabular}{
                >{\centering\arraybackslash}p{2.2cm}     
                >{\arraybackslash}p{1.7cm}             
                >{\centering\arraybackslash}p{1.5cm}   
                >{\centering\arraybackslash}p{1.5cm}   
            }
                \toprule
                Model & Dataset & Joint & KLDA--E \\
                \midrule
                \multirow{4}{*}{\textbf{T5-3B}}  
                            & CLINC     & 96.86$_{\pm0.06}$  & 96.04$_{\pm0.17}$ \\
                            & Banking   & 92.30$_{\pm0.10}$  & 93.77$_{\pm0.05}$ \\
                            & DBpedia   & 94.60$_{\pm0.03}$  & 95.33$_{\pm0.09}$ \\
                            & HWU       & 90.30$_{\pm0.10}$  & 89.31$_{\pm0.27}$ \\
                \midrule
                \multirow{4}{*}{\textbf{Mistral-7B}}  
                            & CLINC     & 97.60$_{\pm0.11}$  & 97.13$_{\pm0.11}$ \\
                            & Banking   & 92.50$_{\pm0.14}$  & 92.53$_{\pm0.12}$ \\
                            & DBpedia   & 95.70$_{\pm0.07}$  & 96.00$_{\pm0.08}$ \\
                            & HWU       & 90.43$_{\pm0.11}$  & 90.02$_{\pm0.09}$ \\
                \bottomrule
            \end{tabular}
        }
    \end{minipage}%
    \hfill
    \begin{minipage}{0.495\textwidth}
        \centering
        \resizebox{\textwidth}{!}{
            \begin{tabular}{
                >{\centering\arraybackslash}p{2.2cm}     
                >{\arraybackslash}p{1.7cm}             
                >{\centering\arraybackslash}p{1.5cm}   
                >{\centering\arraybackslash}p{1.5cm}   
            }
                \toprule
                Model & Dataset & Joint & KLDA--E \\
                \midrule
                \multirow{4}{*}{\textbf{DINOv2-small}}  
                            & CIFAR10     & 97.02$_{\pm0.09}$  & 97.00$_{\pm0.07}$ \\
                            & CIFAR100    & 85.52$_{\pm0.17}$  & 84.21$_{\pm0.08}$ \\
                            & T-ImageNet  & 81.30$_{\pm0.17}$  & 78.67$_{\pm0.08}$ \\
                            & Cars        & 81.88$_{\pm0.23}$  & 81.94$_{\pm0.11}$ \\
                \midrule
                \multirow{4}{*}{\textbf{DINOv2-base}}  
                            & CIFAR10     & 98.54$_{\pm0.06}$  & 98.45$_{\pm0.04}$ \\
                            & CIFAR100    & 90.30$_{\pm0.09}$  & 88.81$_{\pm0.07}$ \\
                            & T-ImageNet  & 86.43$_{\pm0.14}$  & 83.18$_{\pm0.11}$ \\
                            & Cars        & 87.47$_{\pm0.21}$  & 87.45$_{\pm0.14}$ \\
                \bottomrule
            \end{tabular}
        }
    \end{minipage}
    \caption{Final accuracy (\%) of KLDA on both text (left) and image (right) classification datasets, evaluated against joint training with different foundation models.}
    \vspace{-1mm}
    \label{tab:combined}
\end{table*}

\subsection{Comparing KLDA with Other Approaches}
The empirical results, presented in Table \ref{tab:main-results}, reveal the limitations of traditional CIL methods. Even the strongest baseline fails to match the accuracy of NCM, indicating that they struggle to fully utilize the foundation model's representations due to CF.

While NCM provides a robust baseline, its performance remains significantly below that of joint training, indicating that a simple class mean cannot effectively utilize the rich feature representations provided by the foundation model. Using a multivariate Gaussian distribution, LDA achieves improved class separation and higher accuracy. Building on this, KLDA applies kernel methods to enhance feature separability, making the representation space more discriminative. Its ensemble variant (KLDA--E), ultimately closes the gap with Joint training and demonstrates that upper-bound performance can be achieved without modifying the foundation model.

\subsection{KLDA Generalization}
Table \ref{tab:combined} presents KLDA’s performance across both text and image classification datasets, benchmarking it against the joint training upper bound. The experiments span multiple foundation models, confirming that KLDA’s effectiveness is not restricted to a specific model architecture or pre-training paradigm.

For text classification, KLDA--E consistently matches or even exceeds joint training performance, highlighting the effectiveness of leveraging foundation model representations without requiring additional training. In image classification, KLDA achieves results on par with Joint on two datasets, while a small performance gap persists on CIFAR-100 and TinyImageNet. This suggests that vision foundation models may not yet offer representations as robust as those in language models, highlighting a possible area for future improvement.

\section{Justification \& Implications}
\label{sec.justifications}
The key characteristic of the method in \cite{momeni2025continual} is that it does not involve learning feature representations. It \textbf{assumes} that the foundation model provides strong and sufficient feature representations that can be used for continual learning in downstream tasks. The fact that this approach can achieve upper-bound accuracy raises several interesting and potentially important questions:

\textit{\textbf{Is it necessary for continual learning to learn new representations?}} We argue that the assumption of leveraging a foundation model with high-quality feature representations for continual learning is justified and that continual learning does not necessarily need to learn new feature representations. This is based on the following considerations:

\begin{enumerate}
    \item \textbf{Representation Modification Leads to CF.}
    If feature representations are continuously updated as new tasks arrive, forgetting becomes unavoidable. Without access to previous task data, any modification to existing features disrupts representations learned from earlier tasks, leading to performance degradation. While replay mechanisms can help mitigate forgetting, they provide only a limited number of past training samples, which is insufficient for fully adjusting all learned feature representations to achieve joint training accuracy.
    
    \item \textbf{Isolating Task Representations is Problematic.}
    Fully preserving past representations requires isolating task-specific parameters. If each task learns its own features while preventing interference through mechanisms like parameter isolation (e.g., HAT \cite{serra2018overcoming}), forgetting can be avoided \cite{kim2022theoretical,lin2024class}. However, this creates multiple representation spaces within the system. In such a setup, the same input object could be represented differently depending on which task-specific model is used, which is counterintuitive. Ideally, a system should maintain a stable and unique representation of an object.
    
    \item \textbf{Neuroscience Perspective.}
    Having a stable and unique representation of an object in the system aligns well with findings in neuroscience, where research suggests that while representations in early sensory areas, such as the primary visual cortex, may drift, downstream neural circuits extract and maintain stable representations despite these fluctuations \cite{roth2023representation}. Additionally, research in the hippocampus suggests that, despite variability at the single-neuron level, the overall population code maintains stable representations across time, ensuring consistent perception and memory \cite{gonzalez2019persistence}. These findings support the idea that once robust representations are learned, they can remain fixed without requiring constant adaptation.
\end{enumerate}

These perspectives suggest that our assumption of using a foundation model for feature extraction is well-justified. Nevertheless, this does not mean that representation learning is never necessary. If a foundation model lacks sufficiently expressive features for certain classes in a specific domain, additional feature learning may be required. However, it does not mean that the features should be learned in the continual learning process. Instead, it may be more effective or appropriate to fine-tune the foundation model on a large, domain-specific dataset using self-supervised learning before leveraging it for continual learning. This also aligns with human learning: when we are unfamiliar with a domain, distinguishing between different concepts or categories can be difficult. However, after acquiring sufficient knowledge, recognizing and categorizing new information becomes much easier. That said, modern foundation models, trained on diverse large-scale datasets, often provide strong general-purpose feature representations, making additional representation learning unnecessary in many cases.

\textit{\textbf{Do humans learn new features?}}
Neuroscience research suggests that representation learning in the human brain is dependent on developmental stages. Infants exhibit remarkable sensory acuity, and their perceptual skills become fine-tuned to the specific sensory environments they experience \cite{watson2014infant}. However, many issues remain unclear, e.g., whether this fine-tuning is related to how an object is associated with an environment based on the frequency of their co-occurrence or to the change of the representation of the object itself. As a concrete example, when a mother teaches a baby to recognize an apple and a banana in a picture book, does the baby learn the features of the two objects to distinguish them or does the baby already have a good internal representation of each object to recognize their differences and is only learning their names? The latter may be the truth.  

In adults, perception representations become stable, allowing adults to learn new concepts without needing to relearn fundamental features \cite{roth2023representation}.
For example, when people learn to identify a new breed of dog, they do not need to develop entirely new visual features. Instead, they rely on existing representations of fur texture, ear shape, body proportions, and facial structure to differentiate the new breed from others. The learning process primarily involves associating these familiar features with a new category label, rather than constructing new feature representations.

This aligns with the idea that foundation models function similarly to an adult human’s perception--they provide a stable feature space that can be used to recognize new concepts without requiring modifications to the underlying representations. Instead of altering features, learning in such a system is about mapping existing features to new categories. While some degree of feature adaptation may still occur in specialized circumstances, the stability of adult sensory representations suggests that effective learning can often be achieved without modifying the underlying features.

\section{Conclusion}
This paper surveyed the research in CIL leading to the approach that can achieve the upper-bound accuracy of 
joint training on all classes simultaneously. This breakthrough is primarily enabled by pre-trained foundation models, which offer rich representations, eliminating the need for representation learning during the continual learning process. Multiple perspectives are provided to justify the approach of not learning new features in continual learning.
This approach effectively eliminated both CF and ICS problems. Earlier CIL methods suffered from significant accuracy degradation as more tasks are introduced, due to these challenges. This severely limited their real-world applicability, where maintaining high accuracy is crucial. With the new technique, continual learning is now viable for practical applications. 

\appendix
\section*{Ethical Statement}
This paper is a survey of an existing body of literature. There are no ethical issues.

\section*{Acknowledgments}
This work was supported by NSF grants IIS-2229876, IIS-1838770, and CNS-2225427. The computing resources were provided through an NVIDIA GPU grant.

\bibliographystyle{named}
\bibliography{ijcai25}

\end{document}